\documentclass[10pt,twocolumn,letterpaper]{article}

\usepackage{iccv}
\usepackage{times}
\usepackage{epsfig}
\usepackage{graphicx}
\usepackage{amsmath}
\usepackage{amssymb}
\usepackage{subfig}
\usepackage{multirow}
\usepackage{mathabx}
\usepackage{color}

\usepackage[breaklinks=true,bookmarks=false]{hyperref}

\iccvfinalcopy 

\begin{document}

\title{2D-CTC for Scene Text Recognition}

\author{\textsuperscript{1}Zhaoyi Wan\thanks{Authors contribute equally.} ,
\textsuperscript{1}Fengming Xie\footnotemark[\value{footnote}],
\textsuperscript{1}Yibo Liu\footnotemark[\value{footnote}],
\textsuperscript{2}Xiang Bai,
\textsuperscript{1}Cong Yao\\
\textsuperscript{1}Megvii, \textsuperscript{2}Huazhong University of Science and Technology\\
i@wanzy.me, xiefengming@megvii.com, liuyibo@megvii.com,\\
xbai@hust.edu.cn, yaocong2010@gmail.com}

\maketitle

\begin{abstract}
   Scene text recognition has been an important, active research topic in computer vision for years. Previous approaches mainly consider text as 1D signals and cast scene text recognition as a sequence prediction problem, by feat of CTC or attention based encoder-decoder framework, which is originally designed for speech recognition. However, different from speech voices, which are 1D signals, text instances are essentially distributed in 2D image spaces. To adhere to and make use of the 2D nature of text for higher recognition accuracy, we extend the vanilla CTC model to a second dimension, thus creating 2D-CTC. 2D-CTC can adaptively concentrate on most relevant features while excluding the impact from clutters and noises in the background; It can also naturally handle text instances with various forms (horizontal, oriented and curved) while giving more interpretable intermediate predictions. The experiments on standard benchmarks for scene text recognition, such as IIIT-5K, ICDAR 2015, SVP-Perspective, and CUTE80, demonstrate that the proposed 2D-CTC model outperforms state-of-the-art methods on the text of both regular and irregular shapes. Moreover, 2D-CTC exhibits its superiority over prior art on training and testing speed. Our implementation and models of 2D-CTC will be made publicly available soon later.

\end{abstract}

\section{Introduction}

As scene text recognition plays a critical role in many real-world applications, it has consistently drawn much research attention from the computer vision community over decades. However, due to the diversity of scene text (e.g., fonts, colors, and arrangements) , the complexity of backgrounds, as well as tough imaging conditions (e.g., blur, perspective distortion, and partial occlusion), scene text recognition remains an extremely challenging task. 

Early scene text recognition systems detect and classify each character separately, and then join the classified results into sequence predictions~\cite{DBLP:conf/eccv/NovikovaBKL12, photo_ocr, wang2011end, DBLP:conf/cvpr/SmithFL11}. In these methods, character-level annotations are required, which are usually expensive in real-world tasks. Furthermore, there might be mistakes and confusions in character detection and classification, leading to degraded recognition accuracy.

Therefore, inspired by speech recognition, Connectionist Temporal Classification (CTC)~\cite{ctc} is introduced to train image-based sequence recognition end-to-end with unsegmented data and sequence level labels~\cite{shi2017end}. In CTC-based methods~\cite{lee2016recursive,star_net}, the whole sequence labels are used to compute conditional probabilities from sequence predictions directly. Many recent studies~\cite{cheng2017fan,aon,lee2016recursive,aster,scan,visual_attn,bai2018edit} followed the framework where network produces frame-wise predictions and aligns them with labels using CTC or attention mechanism.

\begin{figure}[t]
\begin{center}
   \includegraphics[width=0.8\linewidth]{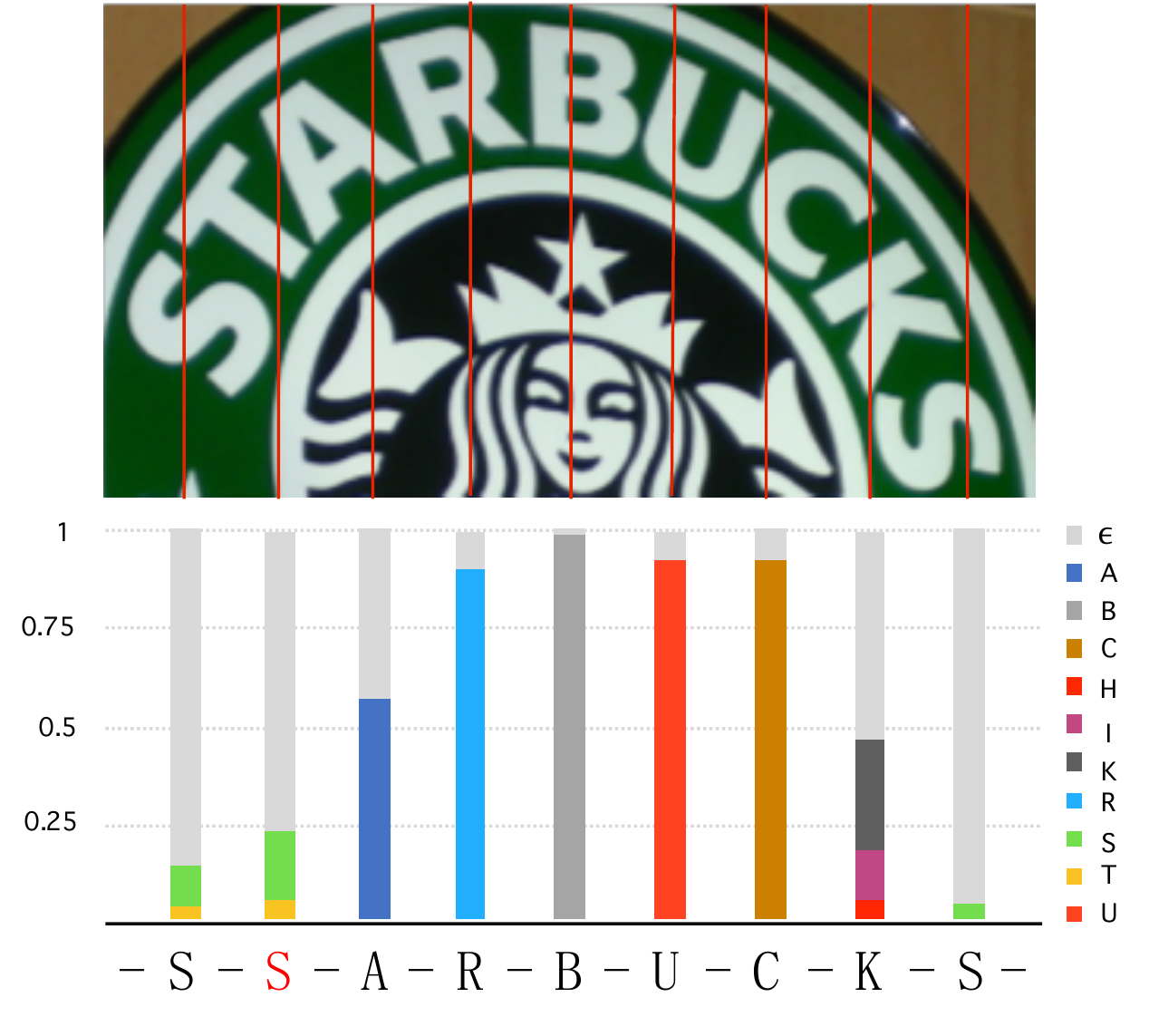}
   \vspace{-2mm}
\end{center}
   \caption{Motivation of 2D-CTC. The prediction process of vanilla CTC is shown above. Color bars represent the probabilities of character classes. At each time step, the character class with the maximum probability is kept. Destruction of spatial information in the vanilla CTC model in the height dimension causes confusion and leads to wrong prediction (T$\rightarrow$S). (Best viewed in color.)}
\label{fig:intro}
\vspace{-4mm}
\end{figure}

\begin{figure*}
\vspace{-2mm}
\begin{center}
     \includegraphics[width=0.8\linewidth]{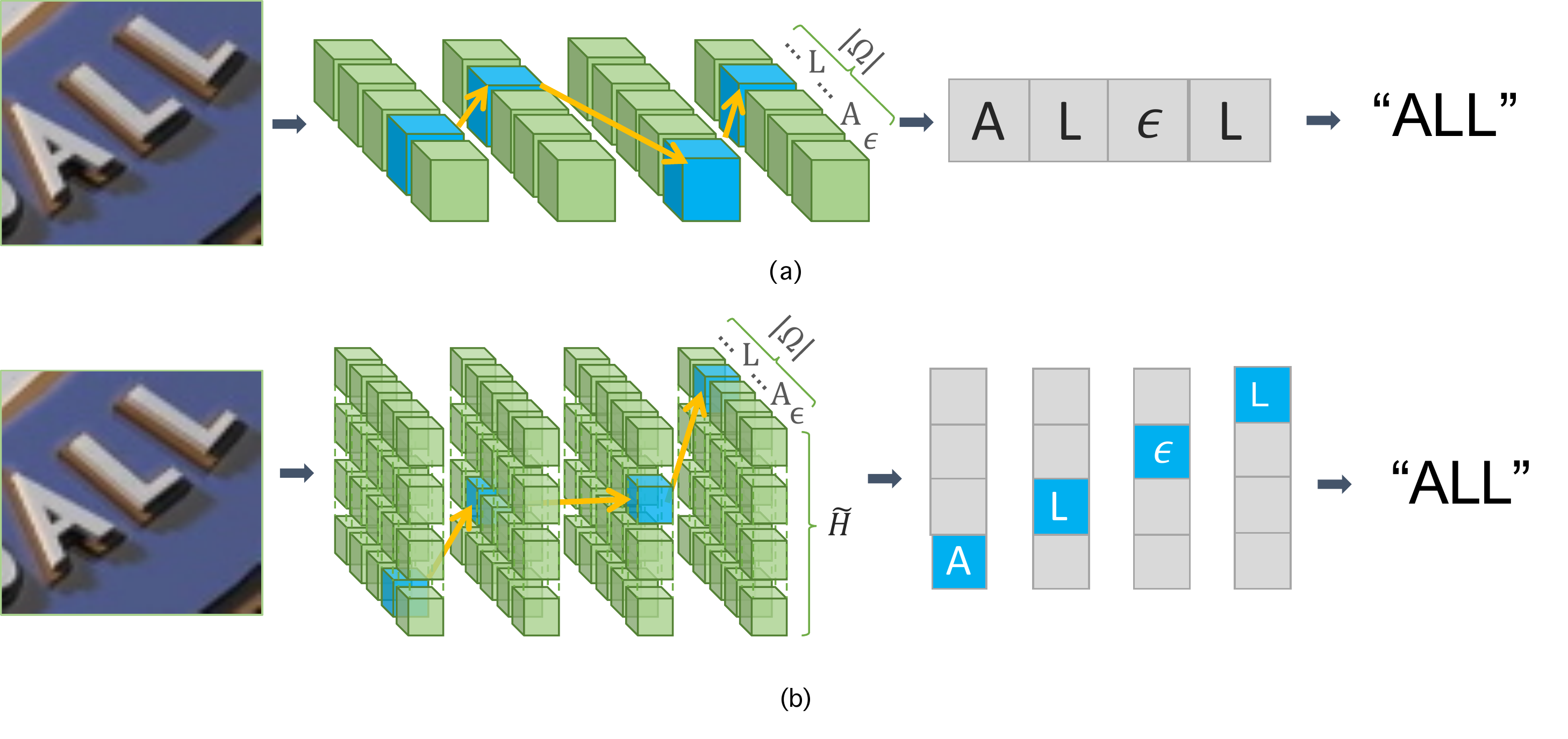}
\end{center}
   \vspace{-6mm}
   \caption{Alignment procedures of vanilla CTC and 2D-CTC. $\Omega$ and $\widetilde{H}$ are the alphabet and the height of probability distribution. (a) The alignment of vanilla CTC. Per-frame probability distributions are predicted and decoded into sequence prediction. (b) The alignment of 2D-CTC. The probability distributions over the height dimension are also computed and utilized. Obviously, compared with vanilla CTC, 2D-CTC performs alignment in a search space with a higher dimension, which endows 2D-CTC with the ability to adaptively focus on relevant features while avoiding the impact of noises.}
\label{fig:long}
\label{fig:onecol}

\end{figure*}

However, vanilla CTC~\cite{ctc} was essentially designed for 1D sequence recognition and can only handle 1D probability distributions. As shown in Fig.\ref{fig:intro}, vanilla CTC-based sequence recognition methods have to collapse 2D features of images into 1D probability distribution at each frame. Due to the significant conflict between 2D text distribution and 1D sequence representation, it may lose crucial information and import extra noises, thus resulting in errors.

To solve this severe limitation of vanilla CTC, we propose a new 2D-CTC (short for two dimensional CTC) formulation to directly compute the conditional probability of labels from 2D distributions.

In vanilla CTC~\cite{ctc}, given a group of probability distributions, it seeks various paths that produce the same target sequence. Summation of the conditional probabilities of all paths conducts the probability of label, which is used to measure the likelihood of label and prediction. As for 2D-CTC, an extra dimension is added for path searching: in addition to time step (length of prediction), probability distributions in the height dimension are preserved. It guarantees that all possible paths over height are taken into consideration, and different path choices over height may still lead to the same target sequence. 

\begin{figure*}[ht]
\centering
\vspace{-2mm}
\includegraphics[width=0.8\linewidth]{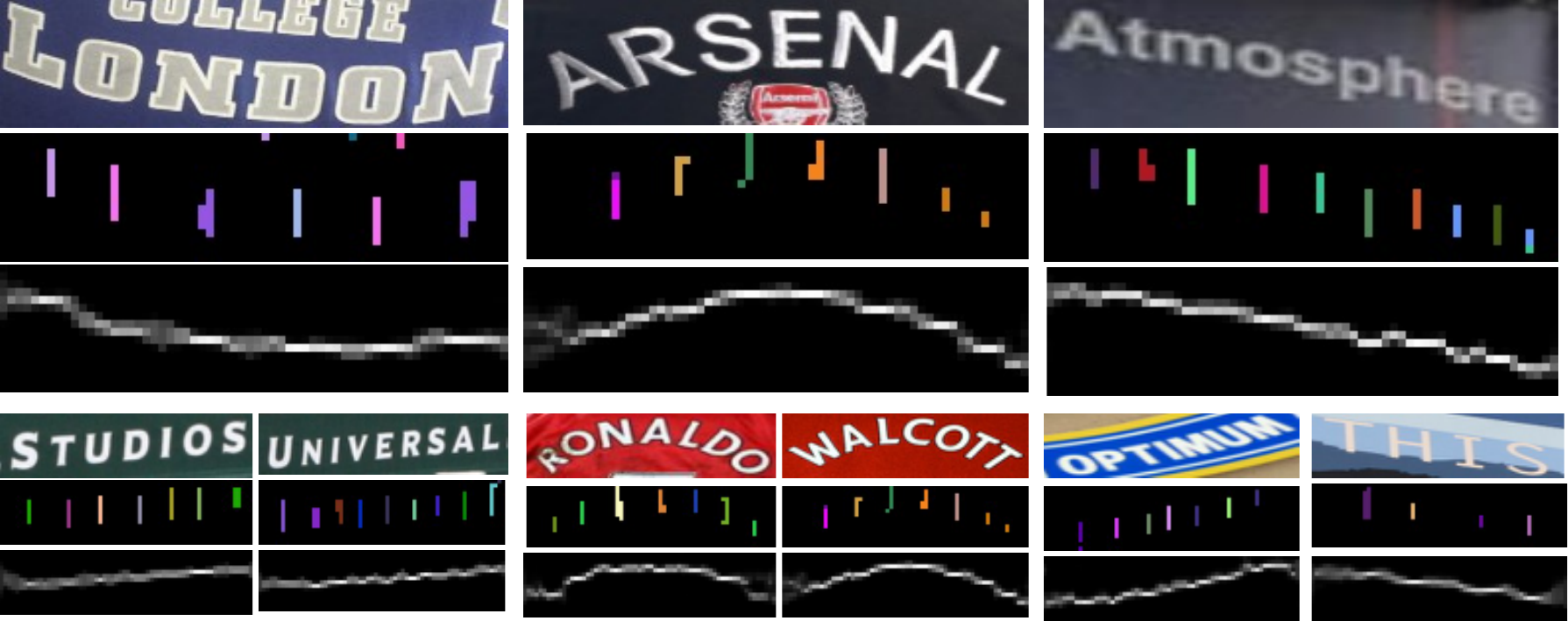}
\vspace{-0.5em}
\caption{Visualization of the intermediate predictions of 2D-CTC. The first rows are input images, the second and last rows are probability distribution maps and path transition maps (See more details in Sec.~\ref{scope}). Note that the predicted distributions and paths, though learned \textit{without} character level annotations, fit well the shapes of the text instances. (Best viewed in color.)}
\label{fig:vis}

\end{figure*}

Without destructing 2D distribution information of characters, more accurate predictions could be achieved. As shown in Fig.~\ref{fig:long}, it is natural for 2D-CTC to recognize curved and irregular text, which is difficult for frame-wise prediction based methods. Furthermore, Fig.~\ref{fig:vis} also shows that our proposed method can detect approximate directions and locations of characters, \textit{even though trained without additional character-level annotations}. This can be used in applications requiring character location information, or weakly supervised character detectors. 

In addition, to reduce the computation burden brought by the dimension increase in 2D-CTC, we devise an effective dynamic programming algorithm(see Sec~\ref{methodology}), which significantly decreases the time complexity of the conditional probability computation.

The contributions of this paper are three-fold: 
(1) We extend the vanilla CTC model by adding another dimension along the height direction to support 2D probability distributions, which can naturally handle various cases such as arbitrarily oriented, curved or rotated text instances. The proposed 2D-CTC model introduces a novel and natural perspective for scene text recognition, where text distributions in 2D space are preserved.
(2) It is demonstrated that the proposed method outperforms the current state-of-the-art method on regular benchmarks, such as IIIT-5K, and archives significant improvement on irregular benchmarks, such as CUTE80 and Total-Text.
(3) We devise an efficient dynamic programming algorithm to compute the conditional probability of a specific label from two or one-dimensional probability distribution. With dynamic programming, the time consumption of 2D-CTC is reduced to an ignorable cost.

\section{Related Work} \label{related_work}

Early scene text recognition methods detect individual characters and recognize each character separately~\cite{DBLP:conf/eccv/NovikovaBKL12, photo_ocr, Jawahar1}. These methods suffer from bad character detection accuracy, which limited recognition performance. 

Inspired by speech recognition, CRNN~\cite{shi2017end} introduce CTC into image-based sequence recognition, which makes it possible to train sequence recognition end-to-end. Following this design, various methods are proposed~\cite{lee2016recursive,star_net}, and achieve significant improvement in performance. However, vanilla CTC is designed for sequence-to-sequence alignment. To fit the formulation, these methods have to collapse 2D image features into 1D distribution, which may lead to loss of relevant information or induction of background noises. Therefore, many of the recent works in scene text recognition have deprecated CTC and turned to other objective functions~\cite{aster, cheng2017fan, Weilin1}.

Another admirable direction of frame-wise prediction alignment is attention-based sequence encoder and decoder framework~\cite{cheng2017fan,aon,lee2016recursive,aster,scan,visual_attn,bai2018edit}. The models focus on one position and predict the corresponding character at each time step, but suffer from the problems of misalignment and attention drift~\cite{cheng2017fan}. Miss-classification at previous steps may lead to drifted attention locations and wrong predictions in successive time steps because of the recursive mechanism. Recent works take attention decoders forward to even better accuracy by suggesting new loss functions\cite{cheng2017fan, bai2018edit} and introducing image rectification modules\cite{aster, Shijian1}. Both of them bring appreciable improvement to attention decoders. However, despite the high accuracy attention decoder has achieved, the considerably lower inference speed is the fundamental factor which has limited its application in real-world text recognition systems. Detailed experiments are presented in Sec \ref{attention-decoder}.

In contrast to the aforementioned methods, Liao~\etal~\cite{ca-fcn} recently propose to utilize instance segmentation to simultaneously predict character locations and recognition results, avoiding the problem of attention misalignment. They also notice the conflict between the 2D image feature and collapsed sequence presentation, and propose a reasonable solution. However, this method requires character-level annotations, and are limited to real-world applications, especially for areas where detailed annotations are hardly available (e.g., handwritten text recognition).

In concern of both accuracy and efficiency, 2D-CTC recognizes text from a 2D perspective similar to~\cite{ca-fcn}, but trained without any character-level annotations. By extending vanilla CTC, 2D-CTC achieves state-of-the-art performance, while retaining the high efficiency of CTC models.
\section{The 2D-CTC Method} \label{methodology}

In this section, we present the details of the proposed 2D-CTC algorithm, which can be applied to image-based sequence recognition problems. Sec~\ref{scope} discusses the output representation of 2D-CTC that allows for the computation of the target probability. Sec~\ref{alignment} describes the procedure where 2D probability distributions are decoded into sequence predictions. Sec~\ref{error measure} construes the dynamic programming algorithm designed for effectively and efficiently computing the 2D-CTC loss.

\subsection{Network Outputs}\label{scope}

The outputs of the network of 2D-CTC can be decomposed into two types of predictions: \textit{probability distribution map}~\cite{bai2018edit} and \textit{path transition map}. 

Firstly we define $\Omega$ as the alphabet. Similar to vanilla CTC, the probability distribution of 2D-CTC is predicted by a softmax output layers with $|\Omega|$ units for $|\Omega|$-class labels. The activation of the first unit at each position represents the probability a `blank', which means no valid label. The following units correspond to the probabilities of the $|\Omega|-1$ character classes, respectively. The shape of probability distribution map is $\widetilde{H}\times \widetilde{W}\times |\Omega|$. $\widetilde{H}$ and $\widetilde{W}$ stand for the spatial size of the prediction maps, which is proportional to the size of the original image.

The distinction between the outputs of vanilla CTC and 2D-CTC is the spatial distribution of predicted probabilities. The output of vanilla CTC models is composed with $\widetilde{T}$ prediction frames, while the probability of 2D-CTC includes prediction at $\widetilde{H}\times \widetilde{W}$ spatial positions. Each predicted vector in the 2D distribution map indicates the classification probabilities of the corresponding spatial location.

To satisfy the definition of 2D-CTC for transition probability, normalization over the height dimension is also required by the 2D-CTC formulation. In practice, an extra prediction, \textit{path transition map}, of shape $\widetilde{H} \times \widetilde{W}$ is produced by another individual softmax layer. Path transition probability is interpreted as the path selection over the vertical direction, which is one of the substantial improvements of 2D-CTC over vanilla CTC. 

The two separate predictions, probability distribution map and path transition map, are used for loss computation and sequence decoding.

\subsection{Prediction Alignment}\label{alignment}

\begin{figure}[!htbp]
\begin{center}
   \includegraphics[width=0.8\linewidth]{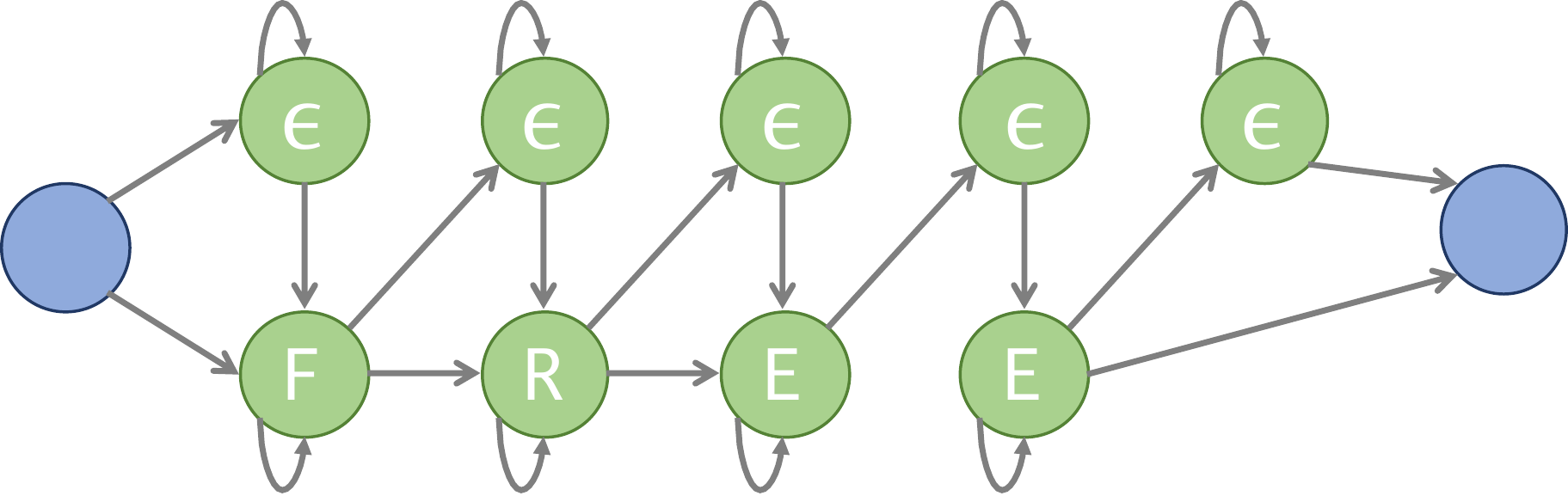}
\end{center}
   
   \caption{Possible alignments of transcription `FREE' in CTC. $\epsilon$ denotes `blank'.}
\label{fig:vanilla-align}
\end{figure}

As discussed in Sec.~\ref{related_work}, vanilla CTC is widely used in sequence recognition. It decodes frame-wise predictions into sequence labels. The key conception of CTC decoding is to skip frames in predictions. To allow for ignoring steps in output paths, vanilla CTC introduced a blank token $\epsilon$. $\epsilon$ implies no valid character and will be removed from the output. Then the predicted sequence can be aligned by ignoring steps where the predicted class is the same with the previous step. A simple example is illustrated in Fig.~\ref{fig:vanilla-align}.

Equipped with CTC algorithm, instead of predicting each corresponding characters in labels, models predict probability distributions of character classes and blank($\epsilon$). The way of alignment-free prediction brings better integration and make it easier for models to converge. However, the classical vanilla CTC formulation still lacks the ability to produce multiple outputs in a single column. Therefore we propose 2D-CTC.

%

2D-CTC inherits the alignment conception of vanilla CTC. In the decoding procedure of 2D-CTC, maximum probabilities over height positions and character classes are reduced to sequence prediction. The alignment of 2D-CTC is illustrated in Fig. \ref{fig:onecol}. Aside from precedent frame alignment of vanilla CTC, 2D-CTC chooses paths over height(see Fig \ref{fig:onecol} (b)). Then the predicted probabilities of character classes on chosen paths will be joint into frame-wise predictions, thus resort the decoding procedure into vanilla CTC alignment.

\subsection{Training the Networks}\label{error measure}

We have discussed the transition from 1D and 2D probability distribution to sequence prediction. To optimize the networks using gradient descent, we now roughly describe the object function of vanilla and our proposed CTC models.

\subsubsection{Vanilla CTC Loss}

As known, the optimizing of CTC models is to maximize the log-likelihood of target labels\cite{ctc}. Formally, the loss function of Vanilla CTC can be summarized as follows:
\begin{equation} \label{1d-loss}
    L_{vanilla} = - \ln{P(Y|X)}
\end{equation}

where X is the predicted probability distribution and Y is the corresponding label. The objective for the probability of $Y$ over $X$ is:
\begin{equation}
    P(Y|X) = \sum_{\pi \in A_{X,Y}} \prod ^{\widetilde{T}}_{t=1} X_{t, \pi_{t}}
\end{equation}
where $A_{X,Y}$ denotes all valid alignments in X over Y, $\widetilde{T}$ is the length of X.

As shown above, $P(Y | X)$ can be computed by summing the probabilities of all valid paths, but the time complexity of straightly adding them together is $O((|\Omega|)^{\widetilde{T}})$. 

Fortunately, dynamic programming provides an efficient way to solve this problem. 

Since the symbol $\epsilon$ before and after any symbol can be handled identically, we make a variant of the label to describe it more clearly:
\begin{equation}
    Y^{*} = [\epsilon, y_{1}, \epsilon, y_{2}, \epsilon, ... , y_{\widetilde{L}},\epsilon], \widetilde{L}=|Y|.
\end{equation}

Given $s \in [1, 2, ..., 2\widetilde{L}+1]$, let $Y^{*}[1:s]$ be the first $s$ symbols of $Y^{*}$, we define $\alpha_{s,t}$ as conditional probability of $Y^{*}[1:s]$ after $t$ time steps. Then $\alpha_{s,t}$ can be computed with divide-and-conquer strategy.

For cases where symbol at $s-1$ can not be ignored, $Y^{*}_{s} = \epsilon$ or $Y^{*}_{s} = Y^{*}_{s-2}$, $\alpha_{s,t}$ can be inferred from the previous step by following equation:

\begin{equation}
    \Hat{\alpha}_{s, t} = (\alpha_{s-1, t-1} + \alpha_{s, t-1}) \cdot X_{t, Y_{s}^{*}}
\end{equation}

For other cases where $Y^{*}_{s-1}$ is $\epsilon$ between different symbols, if $Y^{*}_{s} \neq \epsilon$ and $Y^{*}_{s} \neq Y^{*}_{s-2}$ are satisfied, $Y^{*}_{s-1}$ can be ignored, then the probability of two-symbol shorter state may contribute to $\alpha_{s,t}$:
\begin{equation}
    \Tilde{\alpha}_{s, t} = (\alpha_{s-2, t-1} + \alpha_{s-1, t-1} + \alpha_{s, t-1}) \cdot X_{t, Y_{s}^{*}}
\end{equation}

In combination, state transition equation of dynamic programming for Vanilla CTC can be described as:

\begin{equation}
    \alpha_{s, t} =
    \begin{cases}
        \Hat{\alpha}_{s, t}  & \text{if}\ Y^{*}_{s} = \epsilon\ \text{or}\  Y^{*}_{s} = Y^{*}_{s-2}\text{,}\\
        \Tilde{\alpha}_{s, t} & \text{otherwise}
    \end{cases}
\end{equation}

Finally, $P(Y|X)$ required by loss function can be computed by dynamic programming, with a totally differentiable procedure:
\begin{equation}
    P(Y|X) = P(Y^{*}|X) = \alpha_{2\widetilde{L}, \widetilde{T}} + \alpha_{2\widetilde{L}+1, \widetilde{T}}
\end{equation}

\subsubsection{From Vanilla CTC to 2D-CTC} \label{sec:formulation}

In vanilla CTC, the input probability distribution is limited to sequence. For compatibility of 2D distributions, which are essential to image-based sequence recognition, we extend the vanilla CTC formulation to adapt 2D distributions.

Conceptually, given an input 2D probability distribution map $X^{'}$ and target label $Y$, the target of 2D-CTC loss is also to present the conditional probability $P(Y|X^{'})$ for all valid paths.
This conditional probability is used to measure the diversity between the label and the prediction. 

We follow the assumption where the paths spread from left to right, which is natural for most sequence recognition problems. More formally, given 2D distribution $X^{'}$ with height $\widetilde{H}$ and width $\widetilde{W}$, we define the path transition map $\Psi \in R^{\widetilde{H} \times (\widetilde{W}-1) \times \widetilde{H}}$. $\Psi_{h,w,h^{'}}$ concretely indicates the transition probability of decoding path transfers from $(h,w)$ to $(h^{'}, w+1)$, where $h^{'}, h \in [1,...,\widetilde{H}]$, $w\in [1,...,\widetilde{W}-1]$.

In definition the summation of path transition probabilities from each location equals to one:
\begin{equation}
    \sum_{j=1}^{\widetilde{H}}\Psi_{h,w,j}=1.
\end{equation}

The essential enhancement of 2D-CTC over 1D is the introduction of the dimension along height, which alleviates the problem of losing or mixing information and provides more paths for CTC to decode. This enhancement changes the state transition equations for both two aforementioned partial cases.

Given the same expanded label $Y^{*}$, recursive formula for cases where $s-1$ can not be ignored($Y^{*}_{s} = \epsilon$ or $Y^{*}_{s} = Y^{*}_{s-2}$) comes to:
\begin{equation}
    \begin{split}
    \Hat{\beta}_{s, h, w} = &X'_{h,w,Y^{*}_{s}} \cdot \sum_{j=1}^{\widetilde{H}} (\beta_{s-1, j, w-1}  + \beta_{s, j, w-1}) \cdot \Psi_{j, w-1, h},
    \end{split}
\end{equation}

And for other cases:
\begin{equation}
    \begin{split}
    \Tilde{\beta}_{s, h, w} = &X'_{h,w,Y^{*}_{s}} \cdot \sum_{j=1}^{\widetilde{H}} (\beta_{s-2, j, w-1} + \beta_{s-1, j, w-1} \\
    & + \beta_{s, j, w-1}) \cdot \Psi_{j, w-1, h}.
    \end{split}
\end{equation}

As we can perceive from the equations, conditional probability at each time step is separately computed along height dimension. Paths in height dimension are weighted by corresponding transition probability and summed into distribution presentation.




Thus, the dynamic programming procedure of 2D-CTC can be described as:
\begin{equation}
    \beta_{s, h, w} =
    \begin{cases}
        \Hat{\beta}_{s, h, w}  & \text{if}\ Y^{*}_{s} = \epsilon\ \text{or}\  Y^{*}_{s} = Y^{*}_{s-2},\\
        \Tilde{\beta}_{s, h, w} & \text{otherwise.}
    \end{cases}
\end{equation}

In implementation, we need to define the initialization for the first state of $\beta$:

\begin{equation}
    \beta_{s, h, 1}=
    \begin{cases}
        \Gamma_h\cdot X'_{h, 1, Y^{*}_{s}}& \text{if } s <= 2, \\
        0&  \text{otherwise.}
    \end{cases}
\end{equation}
where $\Gamma \in R^{\widetilde{H}}$,
$\sum_{h=1}^{\widetilde{H}}\Gamma_h=1$.
    
So far as discussed, the conditional probability of 2D-CTC is summarized as:
\begin{equation}
    P(Y|X')=P(Y^{*}|X')=\sum_{j=1}^{\widetilde{H}}(\beta_{2\widetilde{L},j,\widetilde{W}}+\beta_{2\widetilde{L}+1,j,\widetilde{W}}).
\end{equation}

Besides, as an implementation trick, the transition probabilities of the locations in the same column to different locations in the next column can be assumed to be equal:
\begin{equation}
    \small{\Psi_{h,w,i} = \Psi_{h,w,j} \text{ for all } i, j  }\text{, where } i,j \in [1,...,\widetilde{H}].
\end{equation}

It simplifies probability prediction $\Psi \in R^{\widetilde{H} \times (\widetilde{W}-1) \times \widetilde{H}}$ to $\Hat{\Psi} \in R^{(\widetilde{W}-1) \times \widetilde{H}}$. We implemented 2D-CTC with both formulations, considering the assumption or not, the diversity of performances is negligible in our experiments. Thus we recommend using the implementation where transition probability is simplified. The experiments in this work are based on this simplified implementation.
\section{Experiments}

To verify the effectiveness and advantages of our proposed 2D-CTC, we conduct experiments on standard datasets for scene text recognition and compare it with previous methods in this field.

\subsection{Implementation Details}

\begin{figure}[t]
\centering
\includegraphics[width=1.0\linewidth]{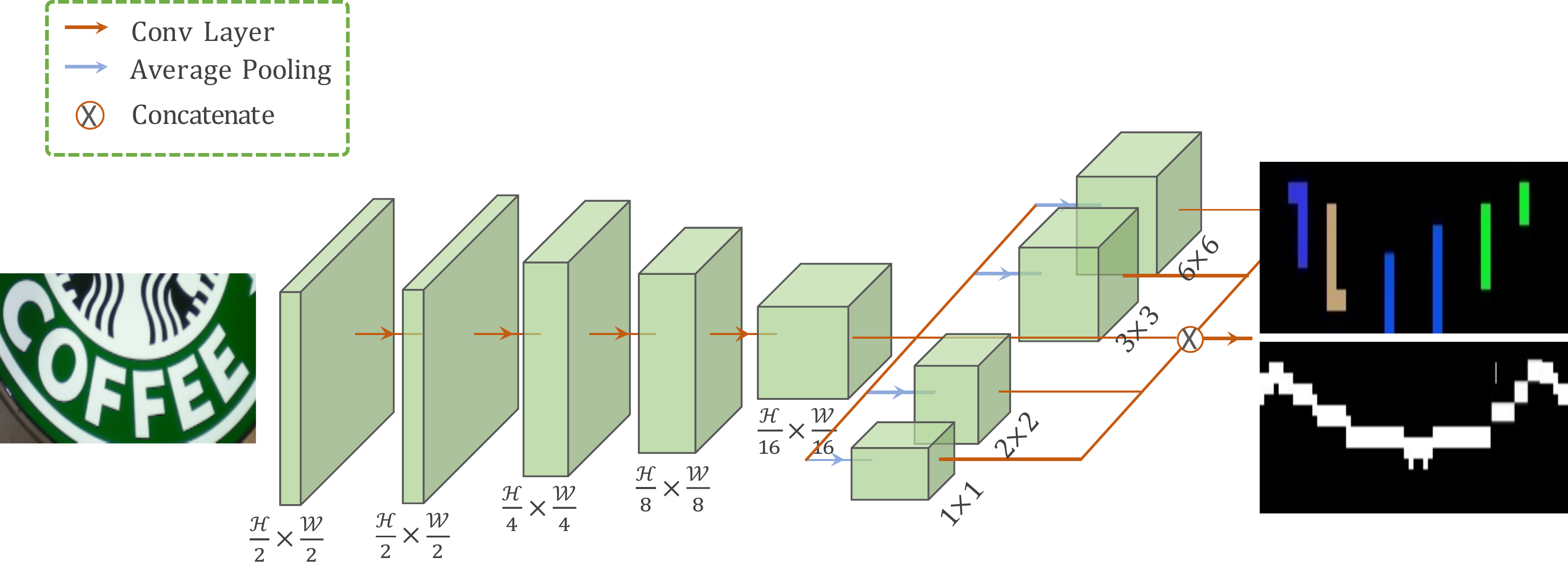}
\caption{Illustration of the network architecture of 2D-CTC. The blocks show the output sizes of convolution layers, where $\mathcal{H}$ and $\mathcal{W}$ indicate the height and width of the input image, respectively. The outputs on the right are probability distribution map and path transition map , both of which are with the spatial shape of $\dfrac{\mathcal{H}}{8}\times \dfrac{\mathcal{W}}{8}$.}
\label{fig:overall}
\vspace{-1em}
\end{figure}

The whole system of 2D-CTC is implemented with PyTorch\cite{paszke2017pytorch}. The overall structure of the network is illustrated in Fig.~\ref{fig:overall}. The base network is identical to the typical PSPNet~\cite{pspnet}. PSPNet is chosen due to its simplicity and excellent performance. The path transition probability map and probability distribution map are produced by a $3\times3$ convolution layer followed by a $1\times1$ convolution layer separately. The input image is resized to 64 pixels in height and 256 pixels in width, for the sake of training in batch. However, as 2D-CTC is a fully-convolutional network, in fact, it can handle images of arbitrary sizes. 

Our models are trained and evaluated with NVIDIA Titan X. The CPU used for efficiency evaluation is 2.40GHz Intel(R) E5-2680. During inference, all images are scaled to 64 pixels in height. Images with an aspect ratio (width over height) smaller than 4 will be resized to fix the width of 256 pixels, while the others will be resized with aspect ratios unchanged. Greedy or beam search can be used to find the path with maximum probability during inference, where beam search brings negligible accuracy improvement. 

We use the Adam\cite{adam} optimizer to train our model with batch size of 256. The learning rate is set to $10^{-3}$ for initialization and is then decayed to $10^{-4}$ and $10^{-5}$.

\subsection{Benchmark Datasets}

Following the settings of recent works, our model is purely trained with synthetic data MJSynth~\cite{mjsynth} and SynthText~\cite{synthetic}, and evaluated on all the standard benchmarks~\textit{without} any further fine-tuning. The benchmarks for evaluation include both regular and irregular text instances. The details of the datasets are described as follows.

~\textbf{IIIT 5K-Words} (IIIT5k) is a dataset released for scene text recognition by~\cite{mishra2012scene}. It consists of 5000 images, including 3000 images for testing and 2000 images for training. Two lexicons are provided for each image, which contains 50 and 1000 words respectively.

~\textbf{Street View Text}~\cite{wang2011end} (SVT) contains road-side images collected from Google Street View. 647 word images were cropped from 249 test images in the dataset. 50-word lexicon associates with each word image is also defined.

~\textbf{ICDAR 2013}~\cite{ic13} (IC13) inherits most of its data from IC03. There are 233 scene images in the dataset that comprise 1,015 cropped word images with ground truths.

~\textbf{ICDAR 2015}~\cite{karatzas2015icdar} (IC15) contains 500 testing images in total. These images are of low quality and have multi-orientation bounding boxes. Following~\cite{cheng2017fan}, we ignore images containing non-alphanumeric characters or irregular text instances, resulting in 1811 cropped text images.

~\textbf{Street View Text Perspective}~\cite{quy2013recognizing} (SVTP) contains 238 street images taken at the same addresses on SVT. 645 word images with a great variety of viewpoints are cropped. It is specifically designed for perspective text recognition.

~\textbf{CUTE80}~\cite{cute} (CUTE) is a dataset focusing on curved text, which contains 80 natural scene images. No lexicon is provided in the dataset.

~\textbf{Total-Text}~\cite{total_text} is a scene text dataset with 2201 cropped images, in which text instances are ranged from slightly to extremely curved.

\subsection{Performances on Standard Benchmarks}

\begin{table*}[!ht]
    \begin{center}
    \vspace{-2mm}
    \begin{tabular}{|c|c|c|c|c|c|c|c|c|c|c|c|}
    \hline
        \multirow{2}{*}{\textbf{Methods}} & \multicolumn{3}{|c|}{IIIT5k} & \multicolumn{2}{|c|}{SVT} & IC13 & IC15 & SVTP & CUTE & TotalText \\
        \cline{2-11}
        & 50 & 1k & 0 & 50 & 0 & 0 & 0 & 0 & 0 & 0 \\
        \hline
        ABBYY~\cite{wang2011end} & 24.3 & - & - & 35.0 & - & - & - & - & - & - \\
        Wang \textit{et al.}~\cite{wang2011end} & - & - & - & 57.0 & - & - & - & - & - & - \\
        Mishra \textit{et al.}~\cite{mishra2012scene} & 64.1 & 57.5 & - & 73.2 & - & - & - & - & - & - \\
        Wang \textit{et al.}~\cite{wang2012end} & - & - & - & 70.0 & - & - & - & - & - & - \\
        Goel \textit{et al.}~\cite{goel2013whole} & - & - & - & 77.3 & - & - & - & - & - & - \\
        Bissacco \textit{et al.}~\cite{bissacco2013photoocr} & - & - & - & - & - & 87.6 & - & - & - & - \\
        Alsharif and Pineau~\cite{alsharif2013end} & - & - & - & 74.3 & - & - & - & - & - & - \\
        Almaz\'{a}n \textit{et al.}~\cite{almazan2014word} & 91.2 & 82.1 & - & 89.2 & - & - & - & - & - & - \\
        Yao \textit{et al.}~\cite{yao2014strokelets} & 80.2 & 69.3 & - & 75.9 & - & - & - & - & - & - \\
        Rodr\'{i}guez-Serrano \textit{et al.}~\cite{rodriguez2015label} & 76.1 & 57.4 & - & 70.0 & - & - & - & - & - & - \\
        Jaderberg \textit{et al.}~\cite{jaderberg2014deep} & - & - & - & 86.1 & - & - & - & - & - & - \\
        Su and Lu~\cite{su2014accurate} & - & - & - & 83.0 & - & - & - & - & - & - \\
        Gordo~\cite{gordo2015supervised} & 93.3 & 86.6 & - & 91.8 & - & - & - & - & - & - \\
        Jaderberg \textit{et al.}~\cite{jaderberg2016reading} & 97.1 & 92.7 & - & 95.4 & 80.7 & 90.8 & - & - & - & - \\
        Jaderberg \textit{et al.}~\cite{jaderberg2014deep} & 95.5 & 89.6 & - & 93.2 & 71.7 & 81.8 & - & - & - & - \\
        Shi \textit{et al.}~\cite{shi2017end} & 97.6 & 94.4 & 78.2 & 96.4 & 80.8 & 86.7 & - & - & - & - \\
        Shi \textit{et al.}~\cite{shi2016robust} & 96.2 & 93.8 & 81.9 & 95.5 & 81.9 & 88.6 & - & 71.8 & 59.2 & - \\
        Lee \textit{et al.}~\cite{lee2016recursive} & 96.8 & 94.4 & 78.4 & 96.3 & 80.7 & 90.0 & - & - & - & - \\
        Yang \textit{et al.}~\cite{yang2017learning} & 97.8 & 96.1 & - & 95.2 & - & - & - & 75.8 & 69.3 & - \\
        Cheng \textit{et al.}~\cite{cheng2017fan} & 99.3 & 97.5 & 87.4 & 97.1 & 85.9 & 93.3 & 70.6 & - & - & - \\
        Cheng \textit{et al.}~\cite{aon} & 99.6 & 98.1 & 87.0 & 96.0 & 82.8 & - & 68.2 & 73.0 & 76.8 & - \\
        Bai \textit{et al.}~\cite{bai2018edit} & 99.5 & 97.9 & 88.3 & 96.6 & 87.5 & \textbf{94.4} & 73.9 & - & - & - \\
        Shi \textit{et al.}~\cite{aster} & 99.6 & $\underline{98.8}$ & $\underline{93.4}$ & $\underline{97.4}$ & $\underline{89.5}$ & 91.8 & \textbf{76.1} & $\underline{78.5}$ & $\underline{79.5}$ & - \\
        \hline
        Vanilla CTC (Baseline) & $\underline{99.8}$ & 97.8 & 92.2 & 95.7 & 86.9 & 91.2 & 69.8 & 75.6 & 77.1 & $\underline{58.3}$ \\
        2D-CTC (Ours) & \textbf{99.8} & \textbf{98.9} & \textbf{94.7} & \textbf{97.9} & \textbf{90.6} & \underline{93.9} & \underline{75.2} & \textbf{79.2} & \textbf{81.3} & \textbf{63.0} \\
    \hline
    \end{tabular}
    \end{center}
    \vspace{-3mm}
    \caption{Results of ours and other methods. ``0", ``50" and ``1k" indicate the size of the lexicons, ``0" means no lexicon. The highest and second highest results are marked with bold fonts and underlines, respectively.}
    \label{tab:performance}
\end{table*}

We first evaluate our proposed 2D-CTC model on the standard benchmarks. A set of intermediate results produced by 2D-CTC are depicted in Fig.~\ref{fig:vis}. The probability distribution maps and path transition maps clearly show that 2D-CTC can adaptively describe the geometric properties of the text instances being recognized, exclude the interference from background clutters and give precise predictions of the classes of the characters and their overall shapes. 

The quantitative results of 2D-CTC, vanilla CTC\footnote{In this work, vanilla CTC has the same base network as 2D-CTC and is trained using the same training data and learning policy. The main difference lies in the formulation of the prediction and loss function.} and other competitors are reported in Tab.~\ref{tab:performance}. As can be seen, 2D-CTC consistently outperforms with large margins vanilla CTC on all the benchmarks (improvements of 2.7\%$\sim$5.4\%). This proves that 2D-CTC, an extension and generalization of vanilla CTC, is effective at recognizing text with various forms (horizontal, oriented and curved).

Compared with previous algorithms, the advantage of 2D-CTC in recognition accuracy is also obvious. Among 10 evaluation settings of the benchmarks, 2D-CTC achieves 8 highest and 2 second highest performances, substantially surpassing all existing text recognition methods (including recent strong competitors such as EP~\cite{bai2018edit} and ASTER~\cite{aster}).

\begin{figure}[!htbp]
\begin{center}
   \includegraphics[width=0.8\linewidth]{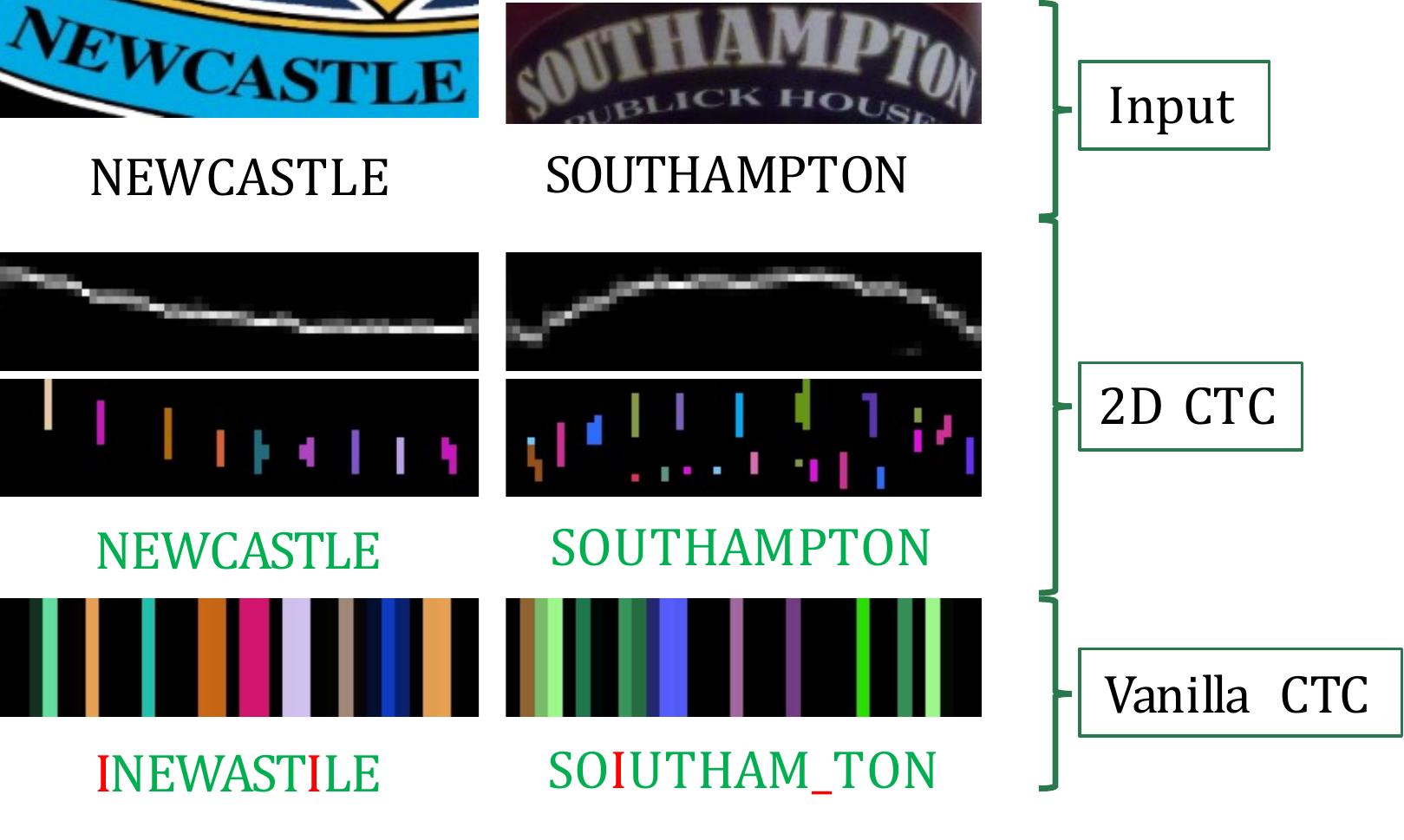}
\end{center}
\vspace{-1em}
   \caption{Visualization of the recognition result of vanilla CTC and 2D-CTC. Note how 2D-CTC ``traces" the true targets (words being recognized) and ``keeps off" the distractors (borders and spurious characters). The heights of the vanilla CTC outputs are expanded to 2D (the same height with that of 2D-CTC) for better visualization. Wrong character predictions are marked with red color and missed characters are indicated with `\_'. }
\label{fig:contrast}
\vspace{-1em}
\end{figure}

Note that 2D-CTC performs particularly well on oriented and curved text, e.g., those from SVTP, CUTE and TotalText. The main reason is that oriented and curved text instances are non-linear objects distributed in 2D spaces, while 2D-CTC is specifically designed to preserve the 2D nature of objects. In contrast to vanilla CTC, in 2D-CTC the collapsing operation along the height direction is eliminated and the optimal alignments are sought in a larger space  ($\widetilde{H}\times \widetilde{W}\times |\Omega|$ instead of~$\widetilde{W}\times |\Omega|$), thus it is possible to better utilize useful information and avoid the negative influence of clutters or noises. This point is also evidenced by the qualitative examples shown in Fig.\ref{fig:contrast}. When clutters or noises appear in the images (notice the borders and small characters in the backgrounds in the first row), vanilla CTC is prone to errors since it simply compressing and employing all features, regardless of the source and quality of the features. However, 2D-CTC handles such cases well, as it can selectively concentrate on features that are relevant and bypass those from clutters or noises. In summary, the ability to model the 2D nature of text is the key to the excellent text recognition performance of 2D-CTC.

\begin{table*}[!ht]
    \begin{center}
    \begin{tabular}{|c|c|c|c|c|c|c|c|c|c|c|}
    \hline
        \multirow{2}{*}{\textbf{Methods}} & \multicolumn{7}{|c|}{\textbf{Performance on Benchmarks (\%$\uparrow$)}} & \multicolumn{3}{|c|}{\textbf{Speed (FPS$\uparrow$)}} \\
        \cline{2-11}
        & IIIT5k & SVT & IC13 & IC15 & SVTP & CUTE & TotalText & Train & Test (GPU) & Test (CPU)\\
        \hline
         Vanilla CTC & 92.2 & 86.9 & 91.2 & 69.8 & 75.6 & 77.1 & 58.3 & 1.42 & 41.63 & 4.55 \\       
        Vanilla CTC + Attention & 91.1 & 88.3 & 91.0 & 70.7 & 76.2 & 78.3 & 60.4 & 1.40 & 39.47 & 4.31 \\
        Attention Decoder~\cite{aster} & 94.4 & 90.5 & 92.6 & 75.8 & 78.3 & 80.4 & 61.3 & 1.15 & 11.35 & 1.13 \\
        2D-CTC & 94.7 & 90.6 & 93.9 & 75.2 & 79.2 & 81.3 & 63.0 & 1.36 & 36.22 & 3.96 \\
    \hline
    \end{tabular}
    \end{center}
    \vspace{-1em}
    \caption{Results of comparative experiments. The speed of models is measured in frames per second (FPS), which indicates how many instances the model can handle in a second. Notably, larger numbers indicate higher efficiency.}
    \label{tab:attention}
    \vspace{-1em}
\end{table*}

\subsection{Comparisons and Discussions}

\subsubsection{Time Cost of Loss Function}

\begin{table}[!t]
    \begin{center}
    \begin{tabular}{|c|c||c|c|}
    \hline
        \multirow{2}{*}{Loss Function}  & \multirow{2}{*}{Device} & \multicolumn{2}{|c|}{Time (ms$\downarrow$)} \\
        \cline{3-4}
        && 1 & 256 \\
    \hline
        \multirow{2}{*}{Vanilla CTC} & CPU & 0.12 & 1.24 \\
         & GPU & 0.23 & 0.24 \\
    \hline
        \multirow{2}{*}{ 2D-CTC } & CPU & 0.2 & 1.98 \\
        & GPU & 0.26 & 0.27 \\
    \hline
    \end{tabular}
    \vspace{-1em}
    \end{center}
    \caption{Computation time costs of vanilla CTC and 2D-CTC on different devices. ``1" and ``256" are the batch sizes.}
    \label{tab:effciency}
    \vspace{-6mm}
\end{table}

The major difference between the loss function of 2D-CTC and that of vanilla CTC is that the former introduces another dimension in the formulation, thus theoretically has higher computation complexity. However, with the proposed dynamic programming algorithm (see Sec.\ref{sec:formulation}), the computation of 2D-CTC loss is constrained to a nearly ignorable cost. As shown in Tab.~\ref{tab:effciency}, both loss functions runs quite fast on CPU and GPU. The actual runtime of the 2D-CTC loss is slightly longer than that of the vanilla CTC loss, which means that 2D-CTC can obtain notable improvement in accuracy, with only a marginal decrease in speed.


\subsubsection{Vanilla CTC with 2D Attention}

Observing the formulation of 2D-CTC, the path transition probabilities can be approximately regarded as a kind of attention, which separates the foreground text from background clutters and noises. To better demonstrate the contribution of the attention-like mechanism, we also introduce a similar attention module into vanilla CTC. The variant ``Vanilla CTC + Attention" has the identical model structure and outputs with 2D-CTC, an extra summation over height dimension is adopted to fit into the formulation of vanilla CTC. The transition probability map can be interpreted as a two-dimensional attention map on prediction. The additional attention module helps to solve images with perspective transformations but does not improve performance on regular text instances. 

As can be observed from Tab.~\ref{tab:attention}, with a different loss formulation, 2D-CTC consistently outperforms ``Vanilla CTC + Attention" on both regular and irregular text benchmarks. This confirms the advantage of the 2D-CTC formulation, which is the essential reason for the improvements in recognition performance.

\subsubsection{CTC \textit{versus} Attention Decoder}\label{attention-decoder}

Besides CTC, attention decoder\cite{lee2016recursive, cheng2017fan, aster} is another widely-used paradigm for sequence prediction, which has been prior art for years. Using RNN and attention mechanism, attention decoder is fed with the global image feature and previous output; then it predicts one character at each time-step. Although remaining obvious drawbacks\cite{cheng2017fan}, methods with attention decoder have proven to outperform CTC-based methods. However, CTC based algorithms are still widely adopted in real-world applications, due to the superiority over attention decoder in running efficiency.

To compare 2D-CTC with attention decoder based methods fairly, we re-implemented ASTER\cite{aster}'s decoder, the state-of-the-art method up to date, with the same base model as 2D-CTC. The evaluated performances and training/test speed are shown in Tab.~\ref{tab:attention}. As can be seen, the proposed 2D-CTC model achieves higher or comparable performances in regular text datasets and outperforms attention decoder in irregular text datasets. 

\begin{figure}[!ht]
\begin{center}
   \includegraphics[width=0.75\linewidth]{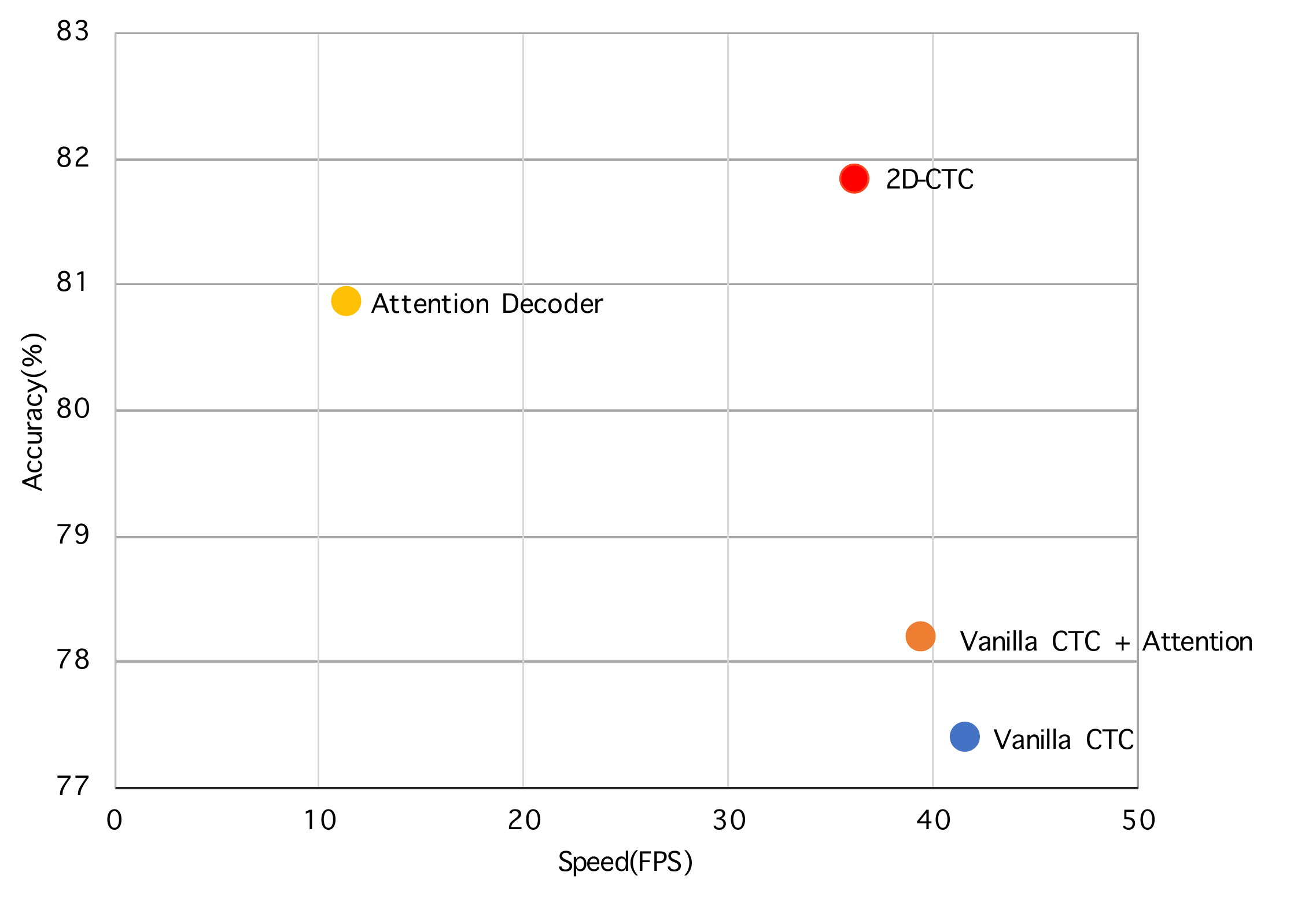}
\end{center}
\vspace{-1.5em}
   \caption{Accuracy and speed of different methods. The plotted accuracy is measured using all images of the aforementioned benchmarks (9953 images in total).}
\label{fig:speed}
\vspace{-1em}
\end{figure}

Fig.~\ref{fig:speed} gives an intuitive comparison of different types of methods in accuracy and speed, where accuracy is measured using the average recognition rate on all the images from the benchmarks and speed is measured using the image number processed per second (in FPS). Attention decoder achieves promising accuracy, but its speed is quite limited, as the built-in recursive modeling in RNNs prevents it from parallelism. In contrast, 2D-CTC outperforms attention decoder in accuracy while running $3\sim 4$ times faster than it (36.22FPS \textit{vs.} 11.35FPS), as 2D-CTC only employs a fully-convolutional network and the formulation allows for efficient inference. This indicates that with the creation of 2D-CTC, the algorithms from the CTC family gain advantages over attention decoder based methods in terms of both accuracy and efficiency.

Nevertheless, the internal capability of modeling the contextual relationships in sequences in attention decoder is actually complementary to the characteristic of the CTC family. For applications that require extra high recognition accuracy, it would be a feasible solution to combine these two types of methodologies, taking the best of the two worlds for higher performance. This direction is worthy of further exploration and investigation.
 
\section{Conclusion}

In this paper, we have presented a novel 2D-CTC model for scene text recognition, which is an extension to the vanilla CTC model. Motivated by the observation that instances of scene text are actually in 2D forms, 2D-CTC is devised and formulated to describe this distinctive property and produce more precise recognition and more explainable intermediate predictions. The qualitative and quantitative results on standard benchmarks confirmed the effectiveness and advantages of 2D-CTC.

{\small
\bibliographystyle{ieee}
\bibliography{egbib}

\begin{thebibliography}{10}\itemsep=-1pt

\bibitem{almazan2014word}
J.~Almaz{\'a}n, A.~Gordo, A.~Forn{\'e}s, and E.~Valveny.
\newblock Word spotting and recognition with embedded attributes.
\newblock {\em IEEE transactions on pattern analysis and machine intelligence},
  36(12):2552--2566, 2014.

\bibitem{alsharif2013end}
O.~Alsharif and J.~Pineau.
\newblock End-to-end text recognition with hybrid hmm maxout models.
\newblock {\em arXiv preprint arXiv:1310.1811}, 2013.

\bibitem{bai2018edit}
F.~Bai, Z.~Cheng, Y.~Niu, S.~Pu, and S.~Zhou.
\newblock Edit probability for scene text recognition.
\newblock In {\em 2018 {IEEE} Conference on Computer Vision and Pattern
  Recognition, {CVPR} 2018, Salt Lake City, UT, USA, June 18-22, 2018}, pages
  1508--1516, 2018.

\bibitem{photo_ocr}
A.~{Bissacco}, M.~{Cummins}, Y.~{Netzer}, and H.~{Neven}.
\newblock Photoocr: Reading text in uncontrolled conditions.
\newblock In {\em 2013 IEEE International Conference on Computer Vision}, pages
  785--792, Dec 2013.

\bibitem{bissacco2013photoocr}
A.~Bissacco, M.~Cummins, Y.~Netzer, and H.~Neven.
\newblock Photoocr: Reading text in uncontrolled conditions.
\newblock In {\em Proceedings of the IEEE International Conference on Computer
  Vision}, pages 785--792, 2013.

\bibitem{cheng2017fan}
Z.~{Cheng}, F.~{Bai}, Y.~{Xu}, G.~{Zheng}, S.~{Pu}, and S.~{Zhou}.
\newblock Focusing attention: Towards accurate text recognition in natural
  images.
\newblock In {\em 2017 IEEE International Conference on Computer Vision
  (ICCV)}, pages 5086--5094, Oct 2017.

\bibitem{aon}
Z.~Cheng, Y.~Xu, F.~Bai, Y.~Niu, S.~Pu, and S.~Zhou.
\newblock {AON:} towards arbitrarily-oriented text recognition.
\newblock In {\em 2018 {IEEE} Conference on Computer Vision and Pattern
  Recognition, {CVPR} 2018, Salt Lake City, UT, USA, June 18-22, 2018}, pages
  5571--5579, 2018.

\bibitem{total_text}
C.~K. Ch'ng and C.~S. Chan.
\newblock Total-text: A comprehensive dataset for scene text detection and
  recognition.
\newblock In {\em Document Analysis and Recognition (ICDAR), 2017 14th IAPR
  International Conference on}, volume~1, pages 935--942. IEEE, 2017.

\bibitem{visual_attn}
S.~K. {Ghosh}, E.~{Valveny}, and A.~D. {Bagdanov}.
\newblock Visual attention models for scene text recognition.
\newblock In {\em 2017 14th IAPR International Conference on Document Analysis
  and Recognition (ICDAR)}, volume~01, pages 943--948, Nov 2017.

\bibitem{goel2013whole}
V.~Goel, A.~Mishra, K.~Alahari, and C.~Jawahar.
\newblock Whole is greater than sum of parts: Recognizing scene text words.
\newblock In {\em Document Analysis and Recognition (ICDAR), 2013 12th
  International Conference on}, pages 398--402. IEEE, 2013.

\bibitem{gordo2015supervised}
A.~Gordo.
\newblock Supervised mid-level features for word image representation.
\newblock In {\em {IEEE} Conference on Computer Vision and Pattern Recognition,
  {CVPR} 2015, Boston, MA, USA, June 7-12, 2015}, pages 2956--2964, 2015.

\bibitem{ctc}
A.~Graves, S.~Fern{\'a}ndez, F.~Gomez, and J.~Schmidhuber.
\newblock Connectionist temporal classification: Labelling unsegmented sequence
  data with recurrent neural networks.
\newblock In {\em Proceedings of the 23rd International Conference on Machine
  learning}, pages 369--376, Pittsburgh, Pennsylvania, USA, 2006. IMLS.

\bibitem{mjsynth}
A.~Gupta, A.~Vedaldi, and A.~Zisserman.
\newblock Synthetic data for text localisation in natural images.
\newblock In {\em Proceedings of the IEEE Conference on Computer Vision and
  Pattern Recognition}, pages 2315--2324, 2016.

\bibitem{Weilin1}
T.~He, Z.~Tian, W.~Huang, C.~Shen, Y.~Qiao, and C.~Sun.
\newblock An end-to-end textspotter with explicit alignment and attention.
\newblock In {\em 2018 {IEEE} Conference on Computer Vision and Pattern
  Recognition, {CVPR} 2018, Salt Lake City, UT, USA, June 18-22, 2018}, pages
  5020--5029, 2018.

\bibitem{jaderberg2014deep}
M.~Jaderberg, K.~Simonyan, A.~Vedaldi, and A.~Zisserman.
\newblock Deep structured output learning for unconstrained text recognition.
\newblock {\em arXiv preprint arXiv:1412.5903}, 2014.

\bibitem{synthetic}
M.~Jaderberg, K.~Simonyan, A.~Vedaldi, and A.~Zisserman.
\newblock Synthetic data and artificial neural networks for natural scene text
  recognition.
\newblock {\em NIPS Deep Learning Workshop}, 2014.

\bibitem{jaderberg2016reading}
M.~Jaderberg, K.~Simonyan, A.~Vedaldi, and A.~Zisserman.
\newblock Reading text in the wild with convolutional neural networks.
\newblock {\em International Journal of Computer Vision}, 116(1):1--20, 2016.

\bibitem{karatzas2015icdar}
D.~Karatzas, L.~Gomez-Bigorda, A.~Nicolaou, S.~Ghosh, A.~Bagdanov, M.~Iwamura,
  J.~Matas, L.~Neumann, V.~R. Chandrasekhar, S.~Lu, et~al.
\newblock Icdar 2015 competition on robust reading.
\newblock In {\em Document Analysis and Recognition (ICDAR), 2015 13th
  International Conference on}, pages 1156--1160. IEEE, 2015.

\bibitem{ic13}
D.~{Karatzas}, F.~{Shafait}, S.~{Uchida}, M.~{Iwamura}, L.~G. i.~{Bigorda},
  S.~R. {Mestre}, J.~{Mas}, D.~F. {Mota}, J.~A. {Almazàn}, and L.~P. {de las
  Heras}.
\newblock Icdar 2013 robust reading competition.
\newblock In {\em 2013 12th International Conference on Document Analysis and
  Recognition}, pages 1484--1493, Aug 2013.

\bibitem{adam}
D.~P. {Kingma} and J.~L. {Ba}.
\newblock Adam: A method for stochastic optimization.
\newblock {\em international conference on learning representations}, 2015.

\bibitem{lee2016recursive}
C.-Y. Lee and S.~Osindero.
\newblock Recursive recurrent nets with attention modeling for ocr in the wild.
\newblock In {\em Proceedings of the IEEE Conference on Computer Vision and
  Pattern Recognition}, pages 2231--2239, 2016.

\bibitem{ca-fcn}
M.~{Liao}, J.~{Zhang}, Z.~{Wan}, F.~{Xie}, J.~{Liang}, P.~{Lyu}, C.~{Yao}, and
  X.~{Bai}.
\newblock Scene text recognition from two-dimensional perspective.
\newblock In {\em Proceedings of the Thirty-Third {AAAI} Conference on
  Artificial Intelligence}, 2019.

\bibitem{star_net}
W.~Liu, C.~Chen, K.-Y.~K. Wong, Z.~Su, and J.~Han.
\newblock Star-net: A spatial attention residue network for scene text
  recognition.
\newblock In {\em BMVC}, volume~2, page~7, 2016.

\bibitem{mishra2012scene}
A.~Mishra, K.~Alahari, and C.~Jawahar.
\newblock Scene text recognition using higher order language priors.
\newblock In {\em BMVC-British Machine Vision Conference}. BMVA, 2012.

\bibitem{Jawahar1}
A.~Mishra, K.~Alahari, and C.~V. Jawahar.
\newblock Enhancing energy minimization framework for scene text recognition
  with top-down cues.
\newblock {\em Computer Vision and Image Understanding}, 145:30--42, 2016.

\bibitem{DBLP:conf/eccv/NovikovaBKL12}
T.~Novikova, O.~Barinova, P.~Kohli, and V.~S. Lempitsky.
\newblock Large-lexicon attribute-consistent text recognition in natural
  images.
\newblock In {\em Computer Vision - {ECCV} 2012 - 12th European Conference on
  Computer Vision, Florence, Italy, October 7-13, 2012, Proceedings, Part
  {VI}}, pages 752--765, 2012.

\bibitem{paszke2017pytorch}
A.~Paszke, S.~Gross, S.~Chintala, and G.~Chanan.
\newblock Pytorch: Tensors and dynamic neural networks in python with strong
  gpu acceleration.
\newblock {\em URL https://github. com/pytorch/pytorch}, 2017.

\bibitem{quy2013recognizing}
T.~Q. {Phan}, P.~{Shivakumara}, S.~{Tian}, and C.~L. {Tan}.
\newblock Recognizing text with perspective distortion in natural scenes.
\newblock In {\em 2013 IEEE International Conference on Computer Vision}, pages
  569--576, Dec 2013.

\bibitem{cute}
A.~Risnumawan, P.~Shivakumara, C.~S. Chan, and C.~L. Tan.
\newblock A robust arbitrary text detection system for natural scene images.
\newblock {\em Expert Systems with Applications}, 41(18):8027 -- 8048, 2014.

\bibitem{rodriguez2015label}
J.~A. Rodriguez-Serrano, A.~Gordo, and F.~Perronnin.
\newblock Label embedding: A frugal baseline for text recognition.
\newblock {\em International Journal of Computer Vision}, 113(3):193--207, Jul
  2015.

\bibitem{shi2017end}
B.~Shi, X.~Bai, and C.~Yao.
\newblock An end-to-end trainable neural network for image-based sequence
  recognition and its application to scene text recognition.
\newblock {\em IEEE transactions on pattern analysis and machine intelligence},
  39(11):2298--2304, 2017.

\bibitem{shi2016robust}
B.~Shi, X.~Wang, P.~Lyu, C.~Yao, and X.~Bai.
\newblock Robust scene text recognition with automatic rectification.
\newblock In {\em Proceedings of the IEEE Conference on Computer Vision and
  Pattern Recognition}, pages 4168--4176, 2016.

\bibitem{aster}
B.~Shi, M.~Yang, X.~Wang, P.~Lyu, C.~Yao, and X.~Bai.
\newblock Aster: An and attentional scene and text recognizer and with flexible
  and rectification.
\newblock In {\em IEEE Transactions on Pattern Analysis and Machine
  Intelligence}, pages 1--1. IEEE, 2018.

\bibitem{DBLP:conf/cvpr/SmithFL11}
D.~L. Smith, J.~Field, and E.~Learned-Miller.
\newblock Enforcing similarity constraints with integer programming for better
  scene text recognition.
\newblock In {\em Proceedings of the 2011 IEEE Conference on Computer Vision
  and Pattern Recognition}, CVPR '11, pages 73--80, Washington, DC, USA, 2011.
  IEEE Computer Society.

\bibitem{su2014accurate}
B.~Su and S.~Lu.
\newblock Accurate scene text recognition based on recurrent neural network.
\newblock In D.~Cremers, I.~Reid, H.~Saito, and M.-H. Yang, editors, {\em
  Computer Vision -- ACCV 2014}, pages 35--48, Cham, 2015. Springer
  International Publishing.

\bibitem{wang2011end}
K.~Wang, B.~Babenko, and S.~Belongie.
\newblock End-to-end scene text recognition.
\newblock In {\em Proceedings of the 2011 International Conference on Computer
  Vision}, ICCV '11, pages 1457--1464, Washington, DC, USA, Nov 2011. IEEE
  Computer Society.

\bibitem{wang2012end}
T.~{Wang}, D.~J. {Wu}, A.~{Coates}, and A.~Y. {Ng}.
\newblock End-to-end text recognition with convolutional neural networks.
\newblock In {\em Proceedings of the 21st International Conference on Pattern
  Recognition (ICPR2012)}, pages 3304--3308, Nov 2012.

\bibitem{scan}
Y.-C. Wu, F.~Yin, X.-Y. Zhang, L.~Liu, and C.-L. Liu.
\newblock Scan: Sliding convolutional attention network for scene text
  recognition.
\newblock {\em arXiv preprint arXiv:1806.00578}, 2018.

\bibitem{yang2017learning}
X.~Yang, D.~He, Z.~Zhou, D.~Kifer, and C.~L. Giles.
\newblock Learning to read irregular text with attention mechanisms.
\newblock In {\em Proceedings of the Twenty-Sixth International Joint
  Conference on Artificial Intelligence, IJCAI-17}, pages 3280--3286, 2017.

\bibitem{yao2014strokelets}
C.~Yao, X.~Bai, B.~Shi, and W.~Liu.
\newblock Strokelets: A learned multi-scale representation for scene text
  recognition.
\newblock In {\em Proceedings of the IEEE Conference on Computer Vision and
  Pattern Recognition}, pages 4042--4049, 2014.

\bibitem{Shijian1}
F.~Zhan and S.~Lu.
\newblock Esir: End-to-end scene text recognition via iterative image
  rectification.
\newblock {\em arXiv preprint arXiv:1812.05824}, 2018.

\bibitem{pspnet}
H.~{Zhao}, J.~{Shi}, X.~{Qi}, X.~{Wang}, and J.~{Jia}.
\newblock Pyramid scene parsing network.
\newblock In {\em 2017 IEEE Conference on Computer Vision and Pattern
  Recognition (CVPR)}, pages 6230--6239, Honolulu, HI, USA, July 2017.

\end{thebibliography}
}

\end{document}